\documentclass[letterpaper]{article} 
\usepackage{aaai24}  
\usepackage{times}  
\usepackage{helvet}  
\usepackage{courier}  
\usepackage[hyphens]{url}  
\usepackage{natbib}  
\usepackage{caption} 
\usepackage{algorithm}
\usepackage{algorithmic}

\usepackage{times}
\usepackage{epsfig}
\usepackage{amsmath}
\usepackage{amssymb}

\usepackage{booktabs}

\usepackage{bm}
\usepackage{algorithm}
\usepackage{algorithmic}
\usepackage{multirow}
\usepackage{color}

\usepackage{float} 
\usepackage{url}            
\usepackage{booktabs}       
\usepackage{amsfonts}       
\usepackage{nicefrac}       
\usepackage{microtype}      
\usepackage{xcolor}         

\newcommand{\eg}{\textit{e}.\textit{g}.}
\usepackage{float}
\usepackage{subfig}
\usepackage{tikz}
\usepackage{comment}
\usepackage{amsmath,amssymb} 
\usepackage{color}
\usepackage{amsmath}
\usepackage{amssymb}
\usepackage[accsupp]{axessibility}  
\usepackage{multirow}
\usepackage{amsfonts}
\usepackage{amsmath}
\usepackage{amsthm}
\usepackage{booktabs}
\usepackage{bm}
\usepackage{algorithm}
\usepackage{algorithmic}

\usepackage{float}

\def\equationautorefname~#1\null{Eq.~(#1)\null}

\usepackage{newfloat}
\usepackage{listings}
\DeclareCaptionStyle{ruled}{labelfont=normalfont,labelsep=colon,strut=off} 
\floatstyle{ruled}
\newfloat{listing}{tb}{lst}{}
\floatname{listing}{Listing}
\title{Ternary Spike: Learning Ternary Spikes for Spiking Neural Networks}
\author{
    xxxx
}

\title{Ternary Spike: Learning Ternary Spikes for Spiking Neural Networks}
\author {
    Yufei Guo\thanks{Equal contribution.}, Yuanpei Chen$^\ast$, Xiaode Liu,  Weihang Peng, Yuhan Zhang,
Xuhui Huang, Zhe Ma\thanks{Corresponding author.}
}
\affiliations {
    Intelligent Science \& Technology Academy of CASIC, China\\
    yfguo@pku.edu.cn, rop477@163.com, mazhe\_thu@163.com
}

\usepackage{bibentry}

\begin{document}

\maketitle

\begin{abstract}
    The Spiking Neural Network (SNN), as one of the biologically inspired neural network infrastructures, has drawn increasing attention recently. It adopts binary spike activations to transmit information, thus the multiplications of activations and weights can be substituted by additions, which brings high energy efficiency. 
    However, in the paper, we theoretically and experimentally prove that the binary spike activation map cannot carry enough information, thus causing information loss and resulting in accuracy decreasing.
    To handle the problem, we propose a ternary spike neuron to transmit information. The ternary spike neuron can also enjoy the event-driven and multiplication-free operation advantages of the binary spike neuron but will boost the information capacity. 
    Furthermore, we also embed a trainable factor in the ternary spike neuron to learn the suitable spike amplitude, thus our SNN will adopt different spike amplitudes along layers, which can better suit the phenomenon that the membrane potential distributions are different along layers.
    To retain the efficiency of the vanilla ternary spike, the trainable ternary spike SNN will be converted to a standard one again via a re-parameterization technique in the inference. 
    Extensive experiments with several popular network structures over static and dynamic datasets show that the ternary spike can consistently outperform state-of-the-art methods.
    Our code is open-sourced at \url{https://github.com/yfguo91/Ternary-Spike}.
\end{abstract}

\section{Introduction}

Nowadays,  Artificial Neural Network (ANN) has been widely used in many fields, \eg, object recognition \cite{2016Deep}, object segmentation \cite{2015U}, object tracking \cite{2016Simple}, mesh smoothing \cite{guo2021new}, etc.
However, to achieve better performance, the network size is evolving to more and more complex \cite{huang2017densely,devlin2018bert}. 
To address this problem, quantization \cite{gong2019differentiable,li2019additive}, pruning \cite{2017Channel}, knowledge distillation \cite{2018Model,2022SelfDistillation}, spiking neural networks (SNNs) \cite{li2021free,2021TrainingXiao,wang2022mesoscopic,2020Online,ren2023spiking,shen2023esl,xu2022hierarchical,xu2021robust}, and so on, have been proposed. 
Especially, SNNs, recognized as one of the next-generation neural networks, provide a unique way to reduce energy consumption by mimicking the information processing of the brain. 
SNNs adopt spikes to transmit information and thus can convert multiplications of weights and activations to additions, enjoying multiplication-free inference. Furthermore, the event-driven-based computation manner shows higher energy efficiency on
neuromorphic hardwares~\cite{2015Darwin,2015TrueNorth,2018Loihi,2019Towards}.

However, SNN's binary spike activation maps suffer the limited information capacity compared to full precious activation maps of the ANN and are unable to carry enough useful information from membrane potentials in the quantization process of the SNN, thus causing information loss and resulting in accuracy decreasing. A detailed proof will be given in Sec.~4.1 in the paper. Meanwhile, we also find that the membrane potential distributions of different layers of an SNN are much different. thus quantizing these membrane potentials to the same spike values is unnatural, which is ignored by the prior work.

\begin{figure*}[t]
	\centering
	\includegraphics[width=0.95\textwidth]{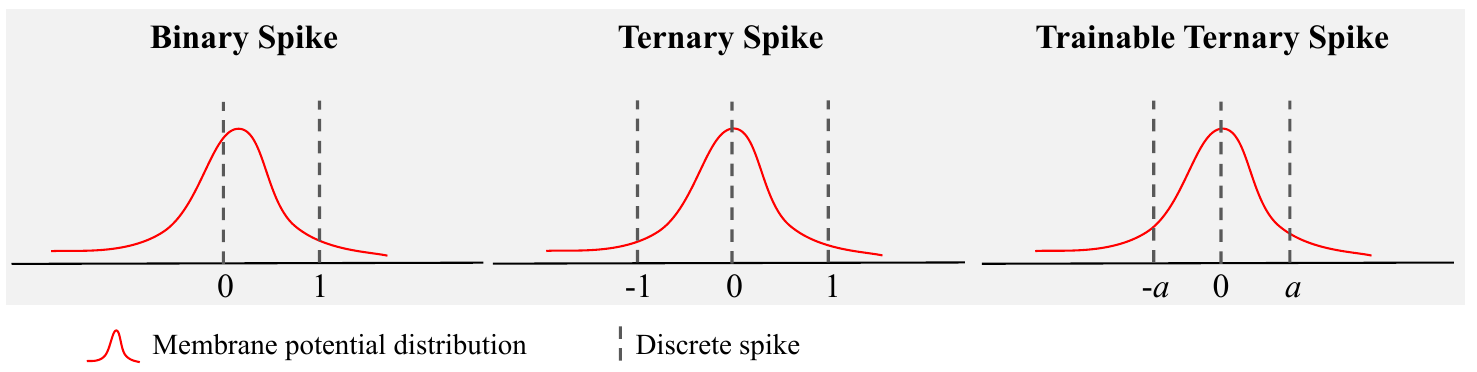} 
	\caption{The difference between our ternary spike neuron and the vanilla binary spike neuron. In binary spike neurons, the membrane potentials will be quantized to $\{0,1\}$ spikes, which will cause severe information loss. In our ternary spike neuron, the membrane potentials will be quantized to $\{-1,0,1\}$ spikes, which will boost the information capacity, while still enjoying the event-driven and multiplication-addition transformation advantages. In trainable spike neurons, the membrane potentials will be quantized to $\{-\alpha,0,\alpha\}$ spikes, where $\alpha$ is learned in the training phase. While in the inference phase, the trainable ternary spike SNN can be converted to a standard ternary spike SNN and retain these advantages of standard SNN still.}
	\label{workflow}
\end{figure*}

To solve these problems, we first propose the ternary spike neuron, called \textbf{Ternary Spike}. Different from the current way using $\{0,1\}$ spikes, it utilizes the $\{-1, 0, 1 \}$ spikes to transmit information. A detailed design and a theoretical proof will be provided in Sec.~4.1 that the \textbf{Ternary Spike} enjoys the greater information capacity than the binary spike but will keep the multiplication-addition transform and event-driven advantages still. Furthermore, we also extend the ternary spike to a learnable ternary spike form, which is not limited to $\{-1, 0, 1 \}$, but $\{-\alpha ,0, \alpha \}$, where $\alpha$ is layer-wise learnable value. In this way, different layers' neurons will fire different magnitudes of spikes in a learning manner, corresponding to different membrane potential distributions. In the inference phase, the $\alpha$ factor can be folded into the weights by a  re-parameterization technique, hence will retain the multiplication-free inference again.
The difference between our ternary spike neuron and the vanilla binary spike neuron is illustrated in Fig.~\ref{workflow}.

 In summary, our main contributions are threefold:
\begin{itemize}
\item We prove that the binary spike activation map cannot carry enough information with theoretical justifications and in-depth experimental analysis and propose the ternary spike neuron, a simple yet effective improved spike neuron for SNNs to increase the information capacity. It still enjoys the multiplication-free and event-driven advantages of the standard SNN. Rather, the ternary spike neuron can be seen as a new paradigm of spike neurons.

\item Furthermore, we also propose a learnable  \textbf{Ternary Spike} version, where the spike magnitude, $\alpha$ can be learned in the training phase. While in the inference phase, the magnitude can be folded into the weights, thus the ternary spikes $\{-\alpha ,0, \alpha \}$ will be transformed to the standard ternary spikes $\{-1, 0, 1 \}$ again and retain the  addition-mostly advantage. 

\item We evaluate our methods on both static (CIFAR-10~\cite{CIFAR-10}, CIFAR-100~\cite{CIFAR-10}, ImageNet~\cite{2009ImageNet}) and spiking (CIFAR10-DVS~\cite{2017CIFAR10}) datasets with widely used backbones. Results show that the SNN trained with the proposed \textbf{Ternary Spike} is highly effective and efficient. \eg, our method reach 70.74\% top-1 accuracy on the ImageNet using ResNet34 with only 4 timesteps,  about 3\% improvement  compared with other state-of-the-art SNN models. 
\end{itemize}

\section{Related Work}
In this section, we briefly overview recent works of SNNs in two aspects: learning methods of SNNs and information loss in SNNs following~\cite{guo2023direct}.

\subsection{Learning Methods of Spiking Neural Networks} 

There are mainly two routers to obtain high-performance deep SNNs. 
The first is converting a well-trained ANN to an SNN, called ANN-SNN conversion~\cite{2020Deep,li2021free,2019Spiking,2020RMP,SpikeConverterFangxin,li2022efficient,2021Constructing,hao2023reducing,hao2023bridging}.
The principle of the ANN-SNN conversion method is to map the parameters from a pre-trained ANN model to its SNN counterpart, based on the matching of the ANN activation values and the SNN average firing rate.
Since the training of an SNN is much more resource and time-consuming than that of the corresponding ANN, this method is one of the widely used  ways to obtain a well-performed SNN.
Nevertheless, There are still several inherent deficiencies in ANN-SNN conversion that are difficult to solve. First, it is limited in the rate-coding scheme and ignores the rich temporal dynamic behaviors from SNNs, thus it cannot handle these neuromorphic datasets well. Second, it usually requires many timesteps to approach the accuracy of pre-trained ANNs. This will increase energy consumption correspondingly, which is contrary to the original intention of SNN's low-power consumption design. Third, the SNN accuracy cannot exceed the ANN accuracy in this paradigm. This will limit the imagination and possibility of the SNN, thus reducing its research values.

Training SNNs directly from scratch is suitable for neuromorphic datasets and can greatly reduce timesteps, even less than 5 in recent work~\cite{deng2022temporal,guo2022real,2021Deep,2018Spatio,2020DIET,2018Direct,2019Surrogate}. Hence this kind of method, as another router of SNN learning has received more research attention recently. 
Besides, the ``hybrid'' learning method~\cite{2020DIET,2021ATandem}, which combines the advantages of ANN-SNN conversion method and direct training method, has also drawn much attention recently. In this work, we focus on improving the performance of the directly training-based SNNs by reducing information loss, which is rarely studied in other works.

\subsection{Information Loss in Spiking Neural Networks} 
There are some works involving reducing the information loss in SNNs\cite{Guo_2022_CVPR,guo2022imloss,Guo2022eccv,guo2023rmploss,wang2023mtsnn}. However, they do not analyze the problem systematically.
In InfLoR-SNN~\cite{Guo2022eccv},  the quantization process of the SNN was thought would cause information loss, and then a membrane potential rectifier that can adjust the membrane potential to a new value closer to quantization spikes than itself before firing activity was proposed. With the same idea of reducing quantization error, the RMP-Loss\cite{guo2023rmploss} was proposed to adjust the membrane potential by a loss function.
IM-Loss~\cite{guo2022imloss} argued that improving the activation information entropy can reduce the information loss, and proposed an information maximization loss function that can maximize the activation information entropy. In RecDis-SNN~\cite{Guo_2022_CVPR}, a loss for membrane potential distribution to explicitly penalize their undesired shifts was proposed. Though the work is not designed for reducing information loss, it still will result in a bimodal membrane
potential distribution, which is proven can mitigate the problem.
In MT-SNN~\cite{wang2023mtsnn}, a multiple threshold (MT) algorithm was proposed at Leak-Integrate-Fire(LIF) neuron to partly recover the
information loss in the quantization process of the SNN.
Nevertheless, these above works all still quantize the membrane potentials
to the binary spikes and ignore the difference in membrane potential distributions along layers. 
In the paper, we present the ternary spike neuron to transmit information with $\{-1, 0, 1 \}$ spikes, which can improve the activation information capacity of the SNN and retain the multiplication-addition transform advantage at the same time. We also improve the proposed ternary spike neuron to a trainable ternary spike neuron, which treats the membrane potential distributions differently. Note that, a ternary spike neuron was also proposed in~\cite{sun2022deep}, however, this neuron transmits information with $\{0, 1, 2 \}$ spikes, and cannot enjoy multiplication-addition transform advantage.

\section{Preliminary}

The brain-inspired spike neuron is the fundamental and special computing unit of SNNs. 
It mimics the behavior of the brain neuron and is described by the interaction of the membrane potential and input current.
We consider the commonly used  Leaky-Integrate-and-Fire (LIF) neuron model in the paper. It can be governed by the following iterative model~\cite{2018Direct},
\begin{equation}\label{eq:u3}
	u^{{\rm before},t}=
		(1-\frac{dt}{\tau}) u^{t-1}+\frac{dt}{\tau}R \cdot I^t,
\end{equation}
\begin{equation}\label{eq:ut}
	u^{t}=
	\left\{
	\begin{array}{lll}
		0, \ \ {\rm if} \; u^{t} \ge V_{\rm th}\\
		u^{{\rm before},t}, \ \ {\rm otherwise} \\
	\end{array},
	\right.
\end{equation}
and 
\begin{equation}\label{eq:ot2}
	o^t=
	\left\{
	\begin{array}{lll}
		1,  \ {\rm if} \;u^{{\rm before},t} \ge V_{\rm th}\\
		0, \ \ {\rm otherwise} \\
	\end{array},
	\right.
\end{equation}
where $u^{{\rm before},t}$ and $u^t$ are the membrane potential before and after the firing processing of the spiking neuron at the $t$-th timestep, $o^t$ is the membrane potential and output spike at the $t$-th timestep, $dt$ and $\tau$ are time constants, the product of $I$ and $R$ denotes the charging voltage by the input current and resistance corresponding to the previous layer signals, and $ V_{{\rm th}}$ is the firing threshold. 
These above equations describe the behaviors of spike neurons in an updating-firing-resetting manner.
When the membrane potential increases up to the firing threshold, the
LIF spike neuron will fire a spike and reset the membrane potential as 0; otherwise, the neuron updates its membrane potential by the addition of the membrane potential after the leakage and the received charging voltage.
Since the factor $(1-\frac{dt}{\tau})$ is a constant, it can still be denoted as $\tau$ and is set as $0.25$ in the paper as \cite{guo2022real,Guo2022eccv}. Considering the SNN is composed of connected spiking neurons with connection coefficients, the input voltage $\frac{dt}{\tau}R \cdot I^t$ can be expanded to the weighted summation of previous layer spike signals as $\sum_jw_jo_{j,\rm pre}^{t}$, where $w_j$ denotes the connected weight of the $j$-th previous layer neuron and current neuron, and $o_{j,\rm pre}^{t}$ indicates the binary spike from the $j$-th previous layer neuron at $t$-th timestep. Further, the iterative LIF model can be simply updated as
\begin{equation}
	u^{t} = \tau u^{t-1}(1-o^{t-1})+\sum_j w_jo_{j,\rm pre}^{t},
	\label{eq:lifu}
\end{equation}
and
\begin{eqnarray}
	 o^{t}=
	\left\{
	\begin{array}{lll}
		1, \ \ {\rm if} \; u^{t} \ge V_{\rm th} \\
		0, \ \ {\rm otherwise}
	\end{array}.
	\right.
	\label{eq:ot-1}
\end{eqnarray}

\section{Methodology}

\subsection{Information Loss in Spiking Neural Networks} 
\label{sec_inf}

Though the binary spike information processing paradigm is highly energy efficient, it will also result in unsatisfactory task performance compared with ANNs. We think one of the most reasons is that the binary spike activation maps cannot carry enough information, thus will cause information loss.
To verify our assumption, we first provide a theoretical analysis here by using the information entropy concept. Given a set, $\textbf{S}$, its representation capability, $\mathcal{R}(\textbf{S})$ can be measured by the information entropy of $\textbf{S}$, as follows
\begin{equation}\label{eq:entroloss}
	\mathcal{R}(\textbf{S}) = \max \mathcal{H}(\textbf{S}) = \max (-\sum_{s \in \textbf S } p_{\textbf S}(s)logp_{\textbf S}(s)),
\end{equation}
where $p_{\textbf S}(s)$ is the probability of a sample, $s$ from $\textbf S$. Then we have the following proposition: 

\noindent\textbf{Proposition 1} \textit{ When $p_{\textbf S}(s_1)=p_{\textbf S}(s_2)=p_{\textbf S}(s_3) \cdots = p_{\textbf S}(s_N)$, $\mathcal{H}(\textbf{S})$ reaches its maximum, $log(N)$.} 

 \noindent Here, $N$ denotes the total number of the samples from $\textbf{S}$. With this conclusion, we can calculate the representation capability of the binary spike feature map and the real-valued membrane potential map. 
Let $\textbf{F}_B\in\mathbb{B}^{C \times H \times W}$ denote as a binary feature map and $\textbf{M}_R\in\mathbb{R}^{C \times H \times W}$ denote as a real-valued membrane potential map.
For a binary spike output $o$, it can be expressed with 1 bit, thus the number of samples from $o$ is $2$. Then, the number of samples from $\textbf{F}_B$ is $2^{(C\times H \times W)}$ and $\mathcal{R}(\textbf{F}_B) = {\bf log} 2^{(C \times H \times W)}=C\times H\times W$. While a real-valued membrane potential needs 32 bits, which consists of $2^{32}$ samples. Hence, $\mathcal{R}(\textbf{M}_R) = {\bf log} 2^{32\times (C\times H\times W)}=32\times C\times H\times W$. It is thus clear that the representation capability of binary spike feature map is much limited and quantizing the real-valued membrane potentials to binary spikes induces excessive information loss. A common consensus is that increasing the timesteps of the SNN can improve the accuracy. This can also be proved by our information theory here. Increasing the timesteps is equivalent to increasing the neuron output spike bits through the temporal dimension, thus increasing the representation capability of the output feature map.

\subsection{Ternary Spike Neuron Model}
\label{tneuron}


The above theoretical analysis shows that improving the spike neuron activation information capacity can increase task performance. And one cannot discard the spike information processing paradigm to increase representation capability, otherwise, the event-driven and addition-only-based energy efficiency will be lost. To boost the information capacity while keeping these advantages, here, we present a ternary LIF spike neuron, given by
\begin{equation}
	u^{t} = \tau u^{t-1}(1-|o^{t-1}|)+\sum_j w_jo_{j,\rm pre}^{t},
	\label{eq:lifternary}
\end{equation}
and 
\begin{equation}\label{oternary}
	o^{t}=
	\left\{
	\begin{array}{lll}
		1, \ \ {\rm if} \; u^{t} \ge V_{\rm th} \\
		-1, \ \ {\rm if} \; u^{t} \le -V_{\rm th} \\
		0, \ \ {\rm otherwise}
	\end{array}.
	\right.
\end{equation}

\noindent\textbf{Representation capacity improvement of ternary spike neuron.} 
We argue that firing ternary spikes can help increase the representation capacity of the SNNs. To verify our assumption, we also resort to the information entropy theory.
Let $\textbf{F}_T\in\mathbb{T}^{C \times H \times W}$ denote as a ternary feature map. The ternary spike $\textbf{F}_T$  consists of $3^{C\times H\times W}$ samples. Hence, $\mathcal{R}(\textbf{F}_T) = {\bf log}_23^{C\times H\times W}$, while $\mathcal{R}(\textbf{F}_B) = C\times H\times W$, according to Eq.~\ref{eq:entroloss}. Obviously, the representation capability of ternary spikes far exceeds that of binary spikes. This indicates that ternary spikes will enhance the information expressiveness of SNNs, which obviously benefits performance improvement. 

\noindent\textbf{Event-driven and addition-only advantages retaining.} 
The SNN's event-driven signal processing characteristic makes it much energy-efficient. In specific, only if the membrane potential exceeds the firing threshold, $V_{\rm th}$, the spiking neuron will present a signal and start subsequent  computations, otherwise, it will keep silent. For the ternary spike neuron, it enjoys the event-driven characteristic too. Rather, only if the membrane potential is greater than $V_{\rm th}$ or less than $-V_{\rm th}$, the ternary spike neuron will be activated to fire the 1 or -1 spikes. 
Multiplication-addition transform is another advantage of SNNs to keep energy-efficient. In a binary spike neuron, when a spike is fired, it will be multiplied by a weight connected to another neuron to transmit information, which can be expressed as
\begin{equation}\label{eo}
	x = 1 \times w.
\end{equation}
Since the spike amplitude is 1, the multiplication can be replaced by an addition operation as
\begin{equation}\label{eo}
	x = 0 + w.
\end{equation}
For a ternary spike neuron, since the spike is 1 or -1, the multiplication will be  
\begin{equation}\label{eo}
	x = 1 \times w, \rm{or} , -1 \times w.
\end{equation}
It can be replaced by an addition operation too as
\begin{equation}\label{eo}
	x = 0 + w, \rm{or} , 0 - w.
\end{equation}
To conclude, the proposed ternary spike neuron will enhance the expression ability of the SNN, at the same time, will retain the event-driven and addition-only advantages as the vanilla SNN too.

\subsection{Trainable Ternary Spike}

\begin{figure*} [tp]
	\centering
	\subfloat[\label{fig:a}]{
		\includegraphics[scale=0.21]{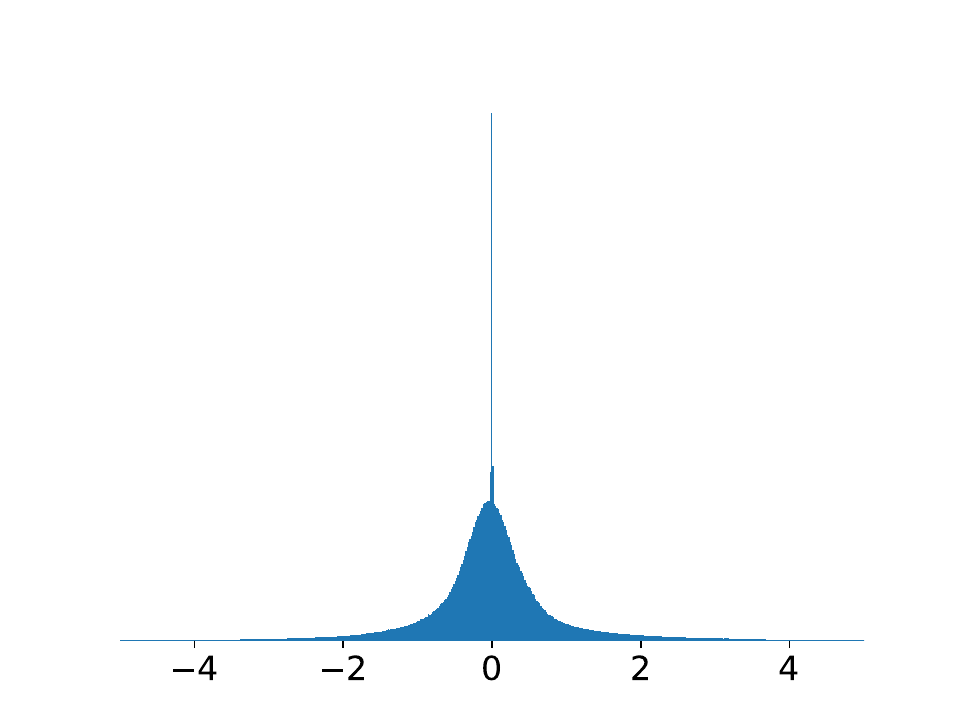}}
	\subfloat[\label{fig:b}]{
		\includegraphics[scale=0.21]{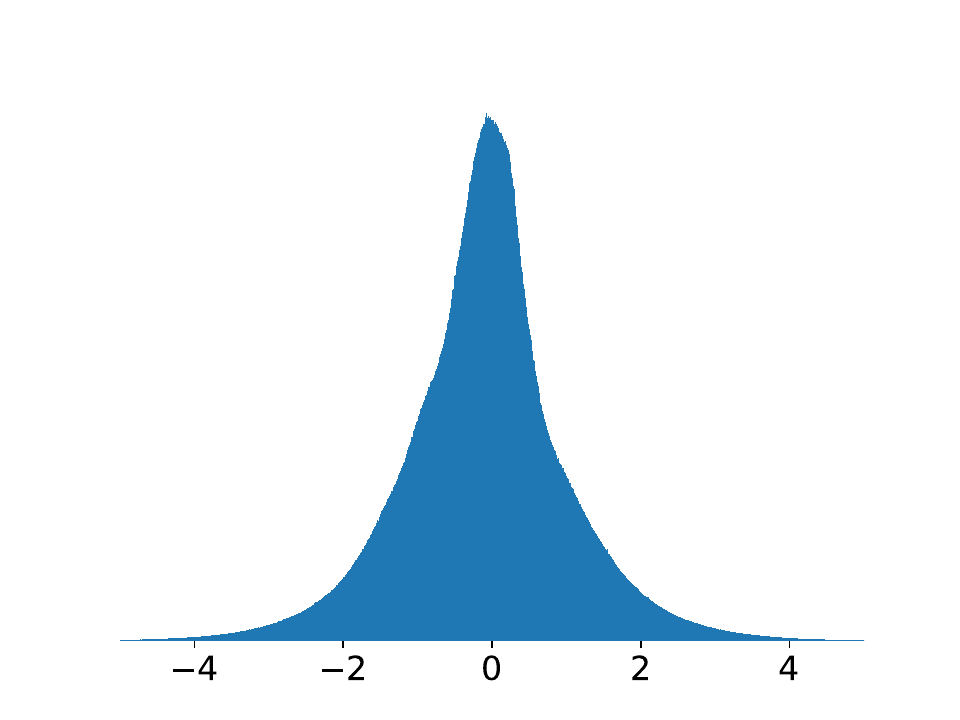}}
	\subfloat[\label{fig:c}]{
		\includegraphics[scale=0.21]{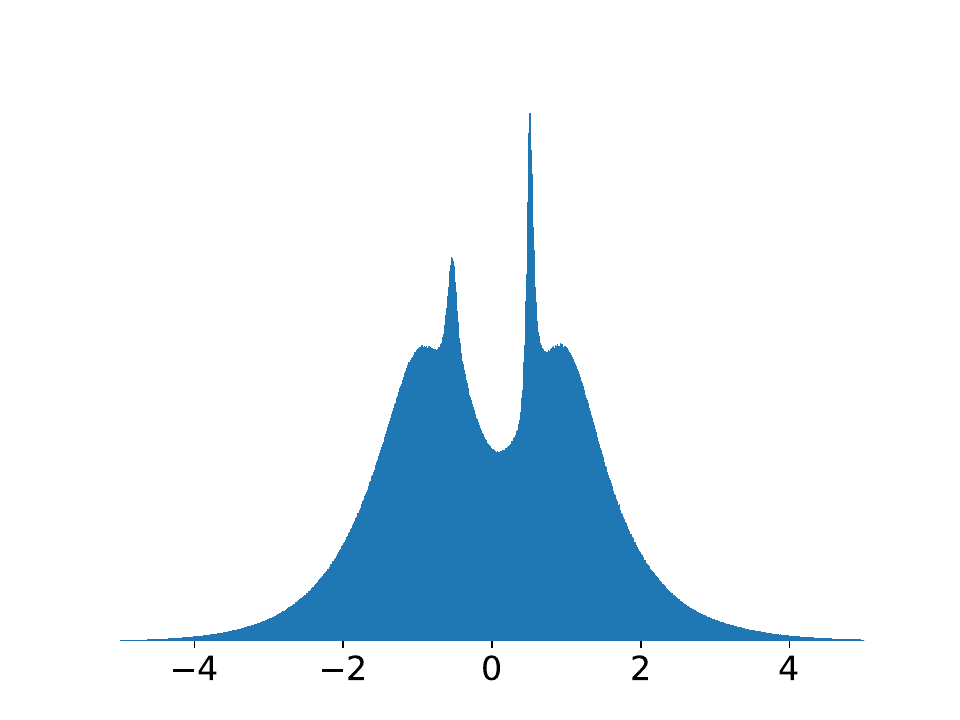}}
    \subfloat[\label{fig:d}]{
		\includegraphics[scale=0.21]{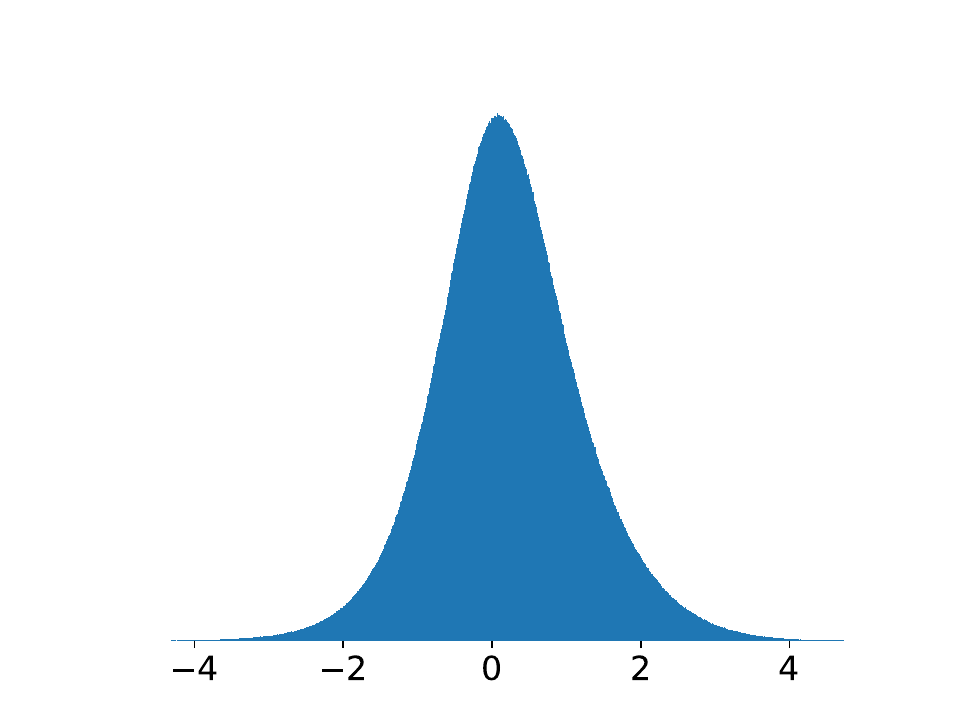}}
	\\
	\subfloat[\label{fig:e}]{
		\includegraphics[scale=0.21]{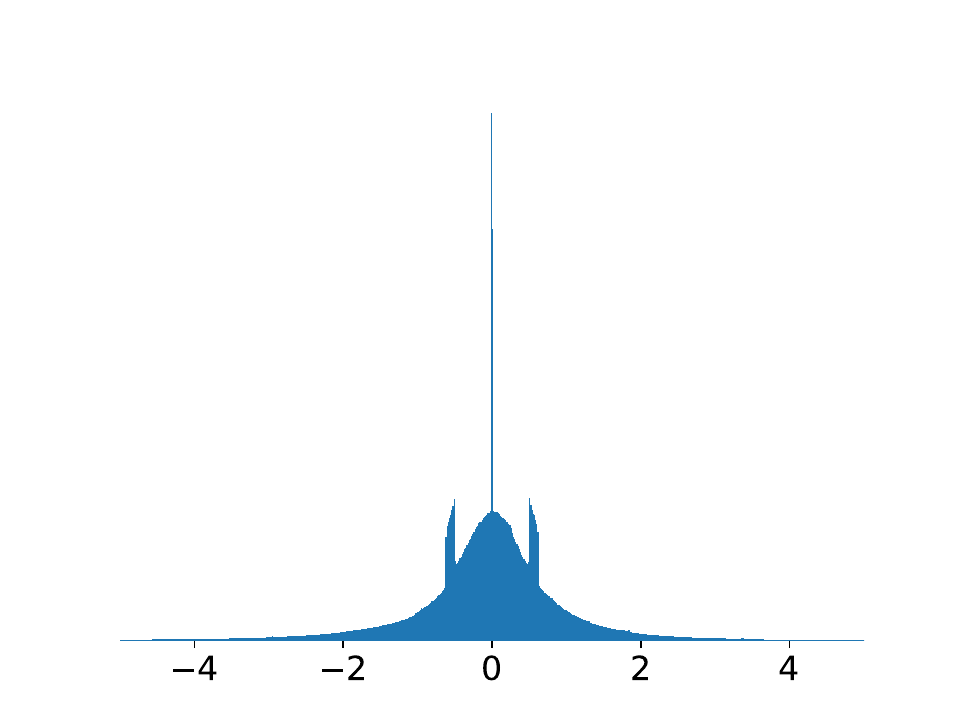} }
	\subfloat[\label{fig:f}]{
		\includegraphics[scale=0.21]{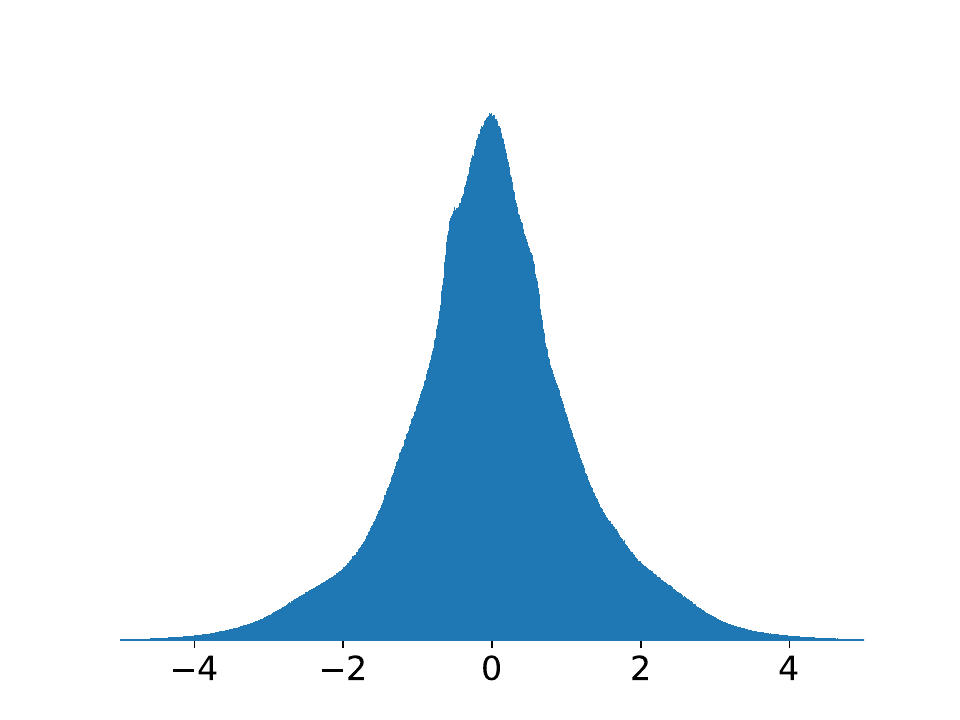}}
	\subfloat[\label{fig:g}]{
		\includegraphics[scale=0.21]{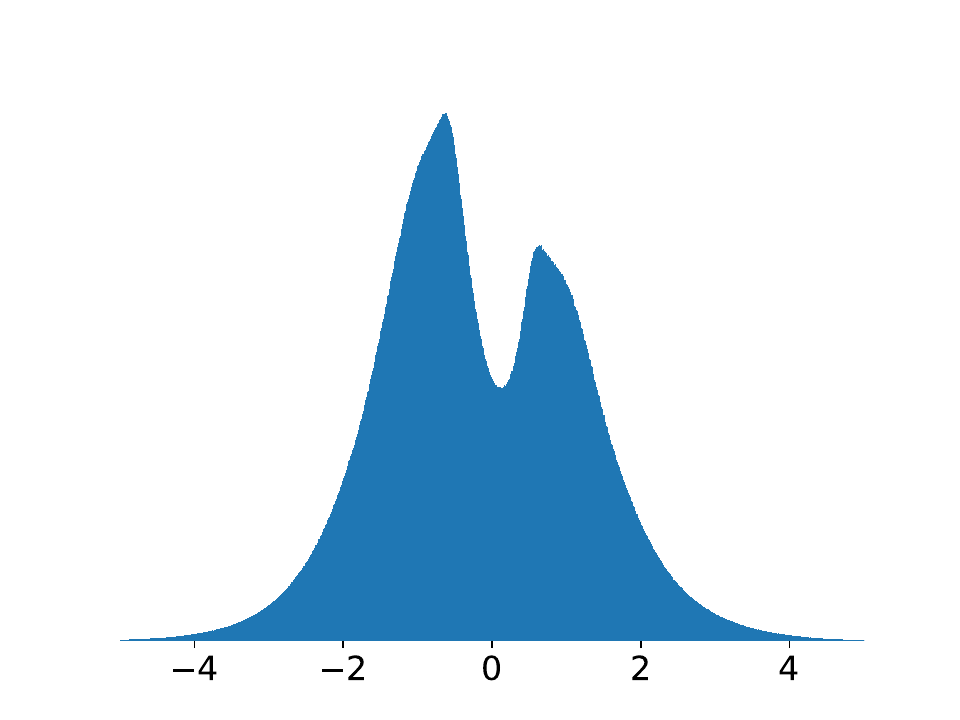}}
	\subfloat[\label{fig:h}]{
		\includegraphics[scale=0.21]{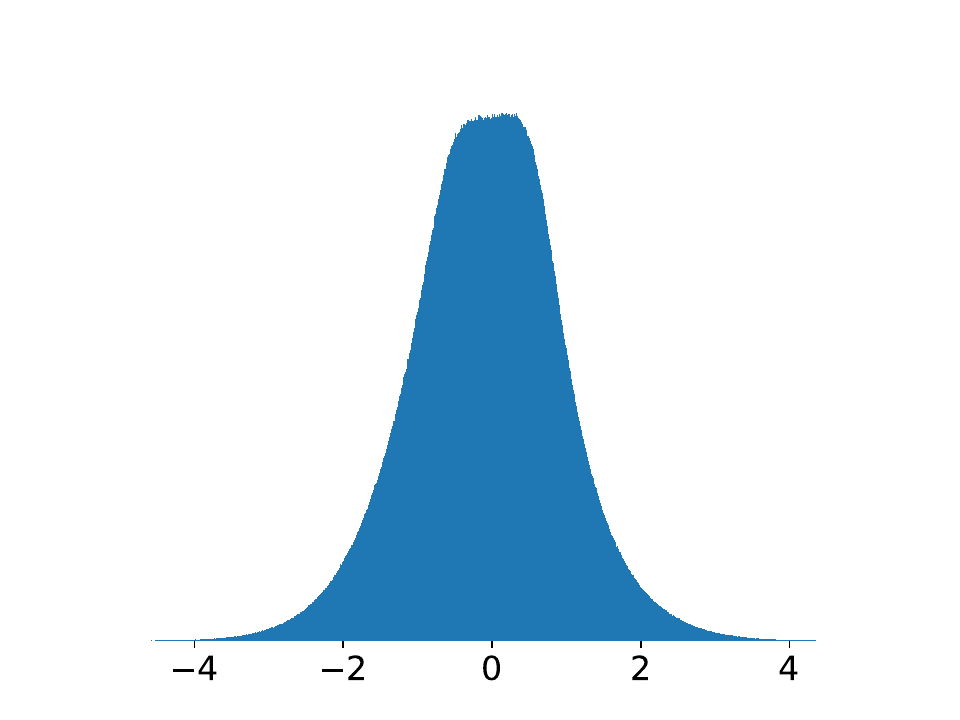}}  
	\caption{The distributions of potential membranes along layers for spiking ResNet20 on CIFAR-10. (a), (b), (c), and (d) show the distributions for the first, second, third, and fourth layers of the network with 1 timestep respectively. (e), (f), (g), and (h) show the distributions for the first, second, third, and fourth layers of the network with 2 timesteps respectively.}
	\label{fig3} 
\end{figure*}

Another problem as we aforementioned is that most prior work quantizes the membrane potentials to the same spike values.
However, in the paper, we find that the membrane potential distributions along layers vary greatly. 
Here, we show the membrane potential distribution differences with some experiments.
We train a spiking ResNet20~\cite{2016Deep} with 1$\&$2 timesteps on the CIFAR-10~\cite{CIFAR-10} dataset and show the different layers' potential membrane distributions in Fig.~\ref{fig3}. It can be seen that the distributions are very different along layers, thus quantizing the different layer’s membrane potentials to the same spike values is unreasonable.
Consequently, we advocate that different layers' membrane potentials should be quantized to different spike values. Then, we present the trainable ternary spike neuron, where the firing spike amplitude can be learned in the training phase given by
\begin{equation}
	u^{t} = \tau u^{t-1}(1-|b^{t-1}|)+\sum_j w_jo_{j,\rm pre}^{t},
	\label{eq:trainedternary}
\end{equation}
and 
\begin{equation}\label{eq:trainedoternary}
	o^{t}=
	\left\{
	\begin{array}{lll}
		1 \cdot a, \ \ {\rm if} \; u^{t} \ge V_{\rm th} \\
		-1 \cdot a, \ \ {\rm if} \; u^{t} \le -V_{\rm th} \\
		0 \cdot a, \ \ {\rm otherwise}
	\end{array},
	\right.
\end{equation}
where $a$ is a trainable factor, $b \in \{-1,0,1\}$, and $o^{t}= a \cdot b^{t}$. With the learnable $a$, the proposed neuron can find a better spike amplitude and treat different layers' firing activity with different strategies. The trainable factor is set in a layer-wise manner in our SNN models, \textit{i.e.}, $\textit{\textbf{a}} \in \mathbb{R}^{1 \times 1 \times 1}$. 

Note that though the trainable factor manner we use in this paper is the same as that used in~\cite{guo2022real}, they are derived from different perspectives. We use the trainable factor to suit the difference of membrane potential distributions in the paper, while the trainable factor is used in ~\cite{guo2022real} to learn unshared convolution kernels. 

However, a new problem will be introduced by the trainable spike that the  multiplication of weight and activation cannot be transformed to addition and the advantage of computation efficiency of SNNs will be lost. To deal with the problem, we follow a training-inference decoupled technique~\cite{guo2022real}, which can convert the different amplitude spikes into the same normalized spike in the inference phase by a re-parameterization technique, thus these advantages of the normalized spike will be retained still. 
Here, we take the convolution layer to introduce the re-parameterization technique.

\noindent\textbf{Re-parameterization technique.}
For a convolution layer, we denote its input feature map and output feature map as $\textbf{F}$ and $\textbf{G}$ respectively. In the convolution layer, the input map will be convolved with a group of convolution kernels to form the output feature map, which can be written as 
\begin{equation}
	\textbf{G}= \textbf{K} * \textbf{F},
	\label{eq:output1}
\end{equation}
where $\textbf{K}$ is the convolution kernel tensor and $(*)$ is the convolution operation. 

For standard ternary SNNs, the input map consists of normalized ternary spikes. While in trainable ternary SNNs, the SNN is trained with real-valued spikes for the purpose of measuring the difference of membrane potential distributions. In this case, the input feature map can be denoted as below according to Eq.~\ref{eq:trainedoternary}
\begin{equation}
	\textbf{F}= a \cdot \textbf{B}.
	\label{eq:decF}
\end{equation}
In inference, we can extract a part of the values from $\textbf{F}$ and fold them into the convolution kernels for the trained SNN. Then a new SNN that can emit normalized ternary spikes will be obtained without changing the values of the output maps. This process can be illustrated as follows:
\begin{equation}
\textbf{G} = \textbf{K} * (a \cdot \textbf{B}) =  (a \cdot \textbf{K}) * \textbf{B} =  \tilde{\textbf{K}} * \textbf{B},
\label{eq1decF}	
\end{equation} 
where $\tilde{\textbf{K}}$ is the transformed convolution kernel tensor. 
The re-parameterization provides us with a solution  to convert a trainable ternary spike-based SNN into an output-invariant normalized ternary spike-based SNN by decoupling the training-time SNN and inference-time SNN.

\begin{table}[tp]
	\centering	
	\caption{Ablation study for the ternary spike on ImageNet.}	
	\label{tab:ab}
	 \setlength{\tabcolsep}{1.0mm}{
	\begin{tabular}{llcc}	
		\toprule
		Architecture & Method & Time-step & Accuracy \\	
		\toprule
		\multirow{6}{*}{ResNet18} & Binary spike & 2 & 58.30\%   \\	
		                                                     &  Ternary spike & 2 & \textbf{65.87\%}   \\
		                                                    &  Trainable ternary spike & 2 & \textbf{66.40\%}   \\                            
		\cline{2-4}
		                             & Binary spike & 4 & 61.07\%   \\	
		                                                     &  Ternary spike & 4 & \textbf{66.90\%}   \\  
		                                                     &  Trainable ternary spike & 4 & \textbf{67.68\%}   \\                              
		\cline{1-4}
		  \multirow{6}{*}{ResNet34} & Binary spike & 2 & 62.81\%   \\	
		                                                     &  Ternary spike & 2 & \textbf{69.48\%}   \\
		                                                     &  Trainable ternary spike & 2 & \textbf{69.51\%}   \\                            
		\cline{2-4}
		                             & Binary spike & 4 & 63.82\%   \\	
		                                                     &  Ternary spike & 4 & \textbf{70.12\%}   \\  
		                                                     &  Trainable ternary spike & 4 & \textbf{70.74\%}   \\     	
		\bottomrule			   		         	            			         	
	\end{tabular}
	}
\end{table}

\section{Experiment}

We performed extensive experiments to evaluate the proposed \textbf{Ternary Spike} and compared it with other recent SoTA methods over several widely used architectures including spiking ResNet20~\cite{2020DIET,2019Going} and ResNet19~\cite{2020Going} on CIFAR-10(100)~\cite{CIFAR-10},  ResNet18~\cite{2021Deep} and ResNet34~\cite{2021Deep} on ImageNet, and ResNet20 and ResNet19 on CIFAR10-DVS~\cite{2017CIFAR10}. 
\begin{table*}[!h]
 \begin{center}
	\caption{Comparison with SoTA methods on CIFAR-10(100).}	
	\label{tab:Comparisoncifar}	
  \resizebox{\textwidth}{!}{
	\begin{tabular}{cllccc}	
		\toprule
		Dataset & Method & Type & Architecture & Timestep & Accuracy \\	
		\toprule
		\multirow{30}{*}{\rotatebox{90}{CIFAR-10}}	
		& SpikeNorm~\cite{2019Going} & ANN2SNN & VGG16 & 2500 & 91.55\%   \\	
		& Hybrid-Train~\cite{2020Enabling} & Hybrid training & VGG16 & 200 & 92.02\%   \\	
		& TSSL-BP~\cite{2020Temporal} & SNN training & CIFARNet & 5 & 91.41\%   \\ 
		& TL~\cite{wu2021tandem} & Tandem Learning & CIFARNet & 8 & 89.04\%   \\   
  		&PTL~\cite{wu2021progressive} & Tandem Learning & VGG11 & 16 &  91.24\%   \\ 
		& PLIF~\cite{2020Incorporating} & SNN training & PLIFNet & 8 & 93.50\%   \\
		& DSR~\cite{meng2022training} & SNN training & ResNet18 &  20 & 95.40\%   \\ 
		& KDSNN~\cite{xu2023constructing} & SNN training & ResNet18 &  4 & 93.41\%   \\ 
  	& Joint A-SNN~\cite{guo2023joint} & SNN training & ResNet18 &  4 & 95.45\%   \\
		\cline{2-6}
		& \multirow{2}{*}{Diet-SNN~\cite{2020DIET}} & \multirow{2}{*}{SNN training} &  \multirow{2}{*}{ResNet20} & 5 & 91.78\%   \\ 
		&  &  &  & 10 & 92.54\%   \\   
		\cline{2-6}
		& \multirow{2}{*}{Dspike~\cite{li2021differentiable}} & \multirow{2}{*}{SNN training} & \multirow{2}{*}{ResNet20} 
		& 2 & 93.13\%   \\
		&  &  &											                                  & 4 & 93.66\%   \\
		\cline{2-6}
		& \multirow{2}{*}{STBP-tdBN~\cite{2020Going}} & \multirow{2}{*}{SNN training} & \multirow{2}{*}{ResNet19} 
		& 2 & 92.34\%   \\
		&  &  &											                                  & 4 & 92.92\%   \\
		\cline{2-6}
		& \multirow{2}{*}{TET~\cite{deng2022temporal}} & \multirow{2}{*}{SNN training} & \multirow{2}{*}{ResNet19} 
		& 2 & 94.16\%   \\
		&  &  &											                                  & 4 & 94.44\%   \\
		\cline{2-6}
		& \multirow{2}{*}{RecDis-SNN~\cite{Guo_2022_CVPR}} & \multirow{2}{*}{SNN training} & \multirow{2}{*}{ResNet19} 
		& 2 & 93.64\%   \\
		&  &  &											                                  & 4 &  95.53\%   \\
		\cline{2-6}
		& \multirow{3}{*}{{RMP-Loss~\cite{guo2023rmploss}}} & \multirow{3}{*}{SNN training} & \multirow{2}{*}{ResNet19} 
		& 2 & {95.31\%} \\
		&  &  &											                                  & 4 & {95.51\%}  \\
		\cline{4-6}	
		&  &  & \multirow{1}{*}{ResNet20} 		                                          & 4 & {91.89\%}  \\
		\cline{2-6}
		& \multirow{4}{*}{\textbf{Ternary Spike}} & \multirow{4}{*}{SNN training} & \multirow{2}{*}{ResNet19} 
		& 1 & \textbf{95.28\%}$\pm 0.10$  \\
		&  &  &											                                  & 2 & \textbf{95.60\%}$\pm 0.09$  \\
		\cline{4-6}	
		&  &  & \multirow{2}{*}{ResNet20} 		                                          & 2 & \textbf{94.29\%}$\pm 0.08$   \\
		&  &  &											                                  & 4 & \textbf{94.46\%}$\pm 0.08$   \\
		\cline{2-6}
		& \multirow{4}{*}{\textbf{Trainable Ternary Spike}} & \multirow{4}{*}{SNN training} & \multirow{2}{*}{ResNet19} 
		& 1 & \textbf{95.58\%}$\pm 0.08$  \\
		&  &  &											                                  & 2 & \textbf{95.80\%}$\pm 0.10$  \\
		\cline{4-6}	
		&  &  & \multirow{2}{*}{ResNet20} 		                                          & 2 & \textbf{94.48\%}$\pm 0.09$   \\
		&  &  &											                                  & 4 & \textbf{94.96\%}$\pm 0.10$   \\  
\hline
		\multirow{17}{*}{\rotatebox{90}{CIFAR-100}}	
		& RMP~\cite{2020RMP} & ANN2SNN & ResNet20 & 2048 & 67.82\%   \\
		& Real Spike~\cite{guo2022real} & SNN training & ResNet20 & 5 & 66.60\%   \\
  	& LTL~\cite{yang2022training} & Tandem Learning & ResNet20 & 31 & 76.08\%   \\ 
		& \multirow{1}{*}{Diet-SNN~\cite{2020DIET}} & \multirow{1}{*}{SNN training} & ResNet20 & 5 & 64.07\%   \\   
		& \multirow{1}{*}{RecDis-SNN~\cite{Guo_2022_CVPR}} & \multirow{1}{*}{SNN training} & ResNet19 & 4 & 74.10\%   \\   
		\cline{2-6}
		& \multirow{2}{*}{Dspike~\cite{li2021differentiable}} & \multirow{2}{*}{SNN training} & \multirow{2}{*}{ResNet20} 
		& 2 & 71.68\%   \\
		&  &  &											                                  & 4 & 73.35\%   \\
		\cline{2-6}
		& \multirow{2}{*}{TET~\cite{deng2022temporal}} & \multirow{2}{*}{SNN training} & \multirow{2}{*}{ResNet19} 
		& 2 & 72.87\%   \\
		&  &  &											                                  & 4 & 74.47\%   \\
		\cline{2-6}

		& \multirow{4}{*}{\textbf{Ternary Spike}} & \multirow{4}{*}{SNN training} & \multirow{2}{*}{ResNet19} 
		& 1 & \textbf{78.13\%}$\pm 0.11$  \\
		&  &  &											                                  & 2 & \textbf{79.66\%}$\pm 0.08$  \\
		\cline{4-6}	
		&  &  & \multirow{2}{*}{ResNet20} 		                                          & 2 & \textbf{73.00\%}$\pm 0.08$   \\
		&  &  &											                                  & 4 & \textbf{73.85\%}$\pm 0.11$   \\
		\cline{2-6}
		& \multirow{4}{*}{\textbf{Trainable Ternary Spike}} & \multirow{4}{*}{SNN training} & \multirow{2}{*}{ResNet19} 
		& 1 & \textbf{78.45\%}$\pm 0.08$  \\
		&  &  &											                                  & 2 & \textbf{80.20\%}$\pm 0.10$  \\
		\cline{4-6}	
		&  &  & \multirow{2}{*}{ResNet20} 		                                          & 2 & \textbf{73.41\%}$\pm 0.12$   \\
		&  &  &											                                  & 4 & \textbf{74.02\%}$\pm 0.08$   \\ 
  
		\bottomrule				         	
	\end{tabular}}	
 \end{center}
\end{table*}
\subsection{Ablation Study}
We first conducted several ablation experiments to verify the effectiveness of the proposed ternary spike neuron model on the ImageNet dataset using ResNet18 and ResNet34  as the backbone respectively under different timesteps. 
The results are shown in Tab.~\ref{tab:ab}.
It can be seen that the highest accuracy for vanilla ResNet18 and ResNet34 are 61.07\% and 63.82\%, similar to existing works.
If we use the ternary spike neuron, the performance would boost to 66.90\% and 70.12\%, which are huge improvements (about 6.0\%).
Moreover, with the trainable ternary spike neuron, the models even further get another performance lift, amounting to 67.68\% and 70.74\% final accuracy, respectively.

\begin{table*}[htbp]
	\centering	
	\caption{Comparison with SoTA methods on ImageNet.}	
	\label{tab:Comparisonimage}	
	\begin{tabular}{llccc}	
		\toprule
		Method & Type & Architecture & Timestep &  Accuracy \\	
		\toprule
				
		STBP-tdBN~\cite{2020Going} &  SNN training & ResNet34 & 6 & 63.72\%   \\ 
		TET~\cite{deng2022temporal} &  SNN training & ResNet34 & 6 & 64.79\%   \\
		RecDis-SNN~\cite{Guo_2022_CVPR} &  SNN training & ResNet34 & 6 & 67.33\%   \\
        OTTT~\cite{xiao2022online} & SNN training & ResNet34 &  6 & 65.15\%   \\ 
        GLIF~\cite{yao2023glif} & SNN training & ResNet34 &  4 & 67.52\%   \\ 
        DSR~\cite{meng2022training} & SNN training & ResNet18 &  50 & 67.74\%   \\
        IM-Loss~\cite{guo2022imloss} & SNN training & ResNet18 &  6 &  67.43\%   \\
		\cline{1-5}
		\multirow{2}{*}{Real Spike~\cite{guo2022real}} & \multirow{2}{*}{SNN training} & {ResNet18} & 4  & {63.68\%}   \\
		 &  &											                             {ResNet34} & 4  & {67.69\%}   \\ 
   \cline{1-5}	
   		\multirow{2}{*}{RMP-Loss~\cite{guo2023rmploss}} & \multirow{2}{*}{SNN training} & {ResNet18} & 4  & {63.03\%}   \\
		 &  &											                             {ResNet34} & 4  & {65.17\%}   \\ 
   \cline{1-5}	
      		\multirow{2}{*}{MPBN~\cite{guo2023membrane}} & \multirow{2}{*}{SNN training} & {ResNet18} & 4  & {63.14\%}   \\
		 &  &											                             {ResNet34} & 4  & {64.71\%}   \\ 
      \cline{1-5}	
      		\multirow{2}{*}{InfLoR-SNN~\cite{Guo2022eccv}} & \multirow{2}{*}{SNN training} & {ResNet18} & 4  & {64.78\%}   \\
		 &  &											                             {ResNet34} & 4  & { 65.54\%}   \\ 
		\cline{1-5}		
		\multirow{2}{*}{SEW ResNet~\cite{2021Deep}} & \multirow{2}{*}{SNN training} & {ResNet18} & 4  & {63.18\%}   \\
		 &  &											                             {ResNet34} & 4  & {67.04\%}   \\ 
		\cline{1-5}
		\multirow{2}{*}{\textbf{Ternary Spike} } & \multirow{2}{*}{SNN training} & {ResNet18} & 4 & \textbf{66.90\%}$\pm 0.19$   \\
		 &  &											                             {ResNet34} & 4& \textbf{70.12\%}$\pm 0.15$   \\
		\cline{1-5}
		\multirow{2}{*}{\textbf{Trainable Ternary Spike} } & \multirow{2}{*}{SNN training} & {ResNet18} & 4 & \textbf{67.68\%}$\pm 0.13$   \\
		 &  &											                             {ResNet34} & 4& \textbf{70.74\%}$\pm 0.11$   \\   
		\bottomrule				         	
	\end{tabular}	
\end{table*}

\begin{table*}[h]
	\centering	
	\caption{Comparison with SoTA methods on CIFAR10-DVS.}	
	\label{tab:Comparisondvs}	
	\begin{tabular}{llccc}	
		\toprule
		 Method & Type & Architecture & Timestep & Accuracy \\	
		\toprule
        DSR~\cite{meng2022training} & SNN training & VGG11 &  20 & 77.27\%   \\
        GLIF~\cite{yao2023glif} & SNN training & 7B-wideNet &  16 & 78.10\%   \\        
		STBP-tdBN~\cite{2020Going} & SNN training & ResNet19 & 10 & 67.80\%   \\ 
		RecDis-SNN~\cite{Guo_2022_CVPR} & SNN training & ResNet19 & 10 & 72.42\%   \\ 
		 \multirow{1}{*}{Real Spike~\cite{guo2022real}} & \multirow{1}{*}{SNN training} & {ResNet19} 
		& 10 & {72.85\%}   \\
		\cline{1-5}
		 \multirow{2}{*}{\textbf{Ternary Spike}} & \multirow{2}{*}{SNN training} & {ResNet19} 
		& 10 & \textbf{78.40\%}$\pm 0.21$   \\
		  &  &											                 {ResNet20} & 10 & \textbf{78.70\%}$\pm 0.17$   \\
		\cline{1-5}
		 \multirow{2}{*}{\textbf{Trainable Ternary Spike}} & \multirow{2}{*}{SNN training} & {ResNet19} 
		& 10 & \textbf{79.80\%}$\pm 0.16$   \\
		  &  &											                 {ResNet20} & 10 & \textbf{79.80\%}$\pm 0.19$   \\    
		\bottomrule				         	
	\end{tabular}	
\end{table*}

\subsection{Comparison with SoTA methods}

In this section, we compared our method with the prior SoTA works.
We report the top-1 accuracy results with the mean accuracy and standard deviation of 3 trials.
We first evaluated our method on CIFAR-10 and CIFAR-100 datasets.
The results are summarized in Tab.~\ref{tab:Comparisoncifar}.
On the CIFAR-10 dataset, the highest accuracy from prior work using ResNet19 and ResNet20 as backbones are 95.51\% and 93.66\% respectively. While our ternary spike achieves 95.60\% and 94.46\% with fewer timesteps. With the trainable ternary spike, Our SNN models can reach a better accuracy.
On the CIFAR-100 dataset, 
our trainable ternary spike using the ResNet19 and ResNet20 with only 2 timesteps outperforms the current best method, TET and RecDis-SNN by about 7\% but with 4 timesteps.
These experimental results clearly show our method's efficiency and effectiveness.

Next, we conducted experiments on the ImageNet dataset, a more complex dataset than CIFAR. The comparison results are shown in Tab.~\ref{tab:Comparisonimage}. There are many SoTA baselines proposed on this dataset recently. For example, RecDis-SNN~\cite{Guo_2022_CVPR}, GLIF~\cite{yao2023glif}, DSR~\cite{meng2022training}, Real Spike~\cite{guo2022real}, and SEW ResNet~\cite{2021Deep} obtain 67.33\%, 67.52\%, 67.74\%, 67.69\%, and 67.04\%, respectively. Yet our method still achieves higher accuracy up to 70.74\%, which is a noteworthy improvement. This shows the effectiveness of our method in handling the large-scale dataset.

Finally, we also ran our SNN on the widely used neuromorphic dataset, CIFAR10-DVS. With ResNet19 and ResNet20 as the backbone, our method scores  79.80\% and 79.80\% accuracy,  close to 80\%, also a huge improvement.

\section{Energy Estimation}
In this section, we measure the hardware energy cost of the binary spike and ternary spike using ResNet20 on CIFAR10 with 2 timesteps for one image inference. Since the first rate-encoding layer does not enjoy the multiplication-free, it will produce the FLOPs (floating point operations). While other layers are calculated by SOPs (synaptic operations). The SOPs are calculated by $s \times T \times A$, where $s$ is the mean sparsity, $T$ is the timestep and $A$ is the addition number in ANN. The sparsity for the SNN with the binary spike is 16.42\%, while for the SNN with the ternary spike is 18.27\%. Note that we see both -1 and 1 as firing. The sign function is only in LIFs, the number of LIFs is limited compared to that of convolution operations. We calculate the energy following~\cite{hu2021spiking}, that one FLOP needs 12.5pJ, one SOP needs 77fJ, and one Sign (using Energy per spike to calculate) needs 3.7pJ. We put the summary of energy cost in Tab.~\ref{tab:Energy}. It can be seen that only 2.11\% extra cost is added for the ternary spike.

\begin{table}[h]
	\centering	
	\caption{Energy estimation.}	
	\label{tab:Energy}	
	\begin{tabular}{llccc}	
		\toprule
		 Method & \#Flops &  \#Sops & \#Sign & Energy \\	
		\toprule
		 Binary Spike & 3.54M & 71.20M & 0.11M & 50.14uJ   \\
        Ternary Spike & 3.54M & 79.21M &  0.23M & 51.20uJ   \\   
		\bottomrule				         	
	\end{tabular}	
\end{table}

\section{Conclusion}
In the paper, we proved that the binary spike activation map of the SNN cannot carry enough information with theoretical justifications and in-depth experimental analysis.
To mitigate the problem, we proposed a ternary spike neuron. This neuron can increase the information capacity greatly and enjoys event-driven and addition-only advantages still.
we also presented an improved ternary spike neuron by embedding a trainable factor in it to learn the suitable spike amplitude.
With learned spike values, this kind of neuron can better suit the different membrane potential distributions along layers.
Furthermore, these different learned spikes can be converted to standard binary spikes in the inference, thus will still enjoy the inherent advantages of SNNs.
We conducted various experiments to verify the effectiveness of our method.

\medskip

\bibliography{aaai24}

\end{document}